\documentclass{article}

\usepackage[final,nonatbib]{nips_2017}
\usepackage[numbers]{natbib}

\usepackage[utf8]{inputenc} 
\usepackage[T1]{fontenc}    
\usepackage{hyperref}       
\usepackage{url}            
\usepackage{booktabs}       
\usepackage{amsfonts}       
\usepackage{nicefrac}       
\usepackage{microtype}      

\usepackage{bbm}
\usepackage{amsmath}
\usepackage{graphicx}

\DeclareMathOperator{\argmax}{arg\,max}

\title{Deep Reinforcement Learning for Sepsis Treatment}

\author{Aniruddh Raghu\\
Cambridge University\\
United Kingdom \\
\texttt{ar753@cam.ac.uk}
\And
Matthieu Komorowski\\
Imperial College London\\
United Kingdom\\
\texttt{m.komorowski14@imperial.ac.uk}
\And
Imran Ahmed\\
Cambridge University\\
United Kingdom\\
\texttt{ia311@cam.ac.uk}
\AND 
Leo Celi\\ 
MIT\\
Cambridge, MA\\
\texttt{lceli@mit.edu}
\And 
Peter Szolovits\\
MIT\\
Cambridge, MA\\
\texttt{psz@mit.edu}
\And 
Marzyeh Ghassemi
\\MIT
\\Cambridge, MA\\
\texttt{mghassem@mit.edu}
}

\begin{document}

\maketitle

\begin{abstract}
 Sepsis is a leading cause of mortality in intensive care units and costs hospitals billions annually. Treating a septic patient is highly challenging, because individual patients respond very differently to medical interventions and there is no universally agreed-upon treatment for sepsis. 
In this work, we propose an approach to deduce treatment policies for septic patients by using continuous state-space models and deep reinforcement learning. Our model learns clinically interpretable treatment policies, similar in important aspects to the treatment policies of physicians. The learned policies could be used to aid intensive care clinicians in medical decision making and improve the likelihood of patient survival. 
\end{abstract}

\section{Introduction} 
Sepsis (severe infections with organ failure) is a dangerous condition that costs hospitals billions of pounds in the UK alone \citep{sepsiscost}, and is a leading cause of patient mortality \citep{sepsismortality}. The clinicians' task of deciding treatment type and dosage for individual patients is highly challenging. Besides antibiotics and infection source control, severe infections are managed by administering intravenous fluids and vasopressors. Various fluids and vasopressor treatment strategies have been shown to lead to extreme variations in patient mortality, which demonstrates how critical these decisions are \citep{waechter2014interaction}. While international efforts attempt to provide general guidance for treating sepsis, physicians at the bedside still lack efficient tools to provide individualized real-term decision support \citep{rhodes2017surviving}. 

In this work, we propose a data-driven approach to discover sepsis treatment strategies, developing on earlier research \citep{komorowski}. We use deep reinforcement learning (RL) algorithms to identify how best to treat septic patients in the intensive care unit (ICU) to improve their chances of survival. We prefer RL for sepsis treatment over supervised learning, because the ground truth of a ``good'' treatment strategy is unclear in medical literature \citep{demise-egdt}. Importantly, RL algorithms also allow us to infer optimal strategies from training examples that do not represent optimal behavior. We focus on continuous state-space modeling, representing a patient's physiological state at a point in time as a continuous vector (using physiological data from the ICU) and find suitable actions with Deep-Q Learning \citep{atari}. 
Our contributions include: deep reinforcement learning models with continuous-state spaces and shaped reward functions; and learned treatment policies that could improve patient outcomes and reduce patient mortality.

\section{Background and related work}
In reinforcement learning, time-varying state spaces are often modeled with Markov Decision Processes (MDP). At every timestep $t$, an agent observes the current state of the environment $s_t$, takes an action $a_t$, receives a reward $r_t$, and transitions to a new state $s_{t+1}$. The agent selects actions that maximize its expected discounted future reward, or \emph{return}, defined as $R_t = \sum_{t'=t}^{T} \gamma^{t'-t}r_{t'}$, where $\gamma$ captures the tradeoff between immediate and future rewards, and $T$ is the terminal timestep. The optimal action value function, ${Q^{*}(s,a) = \max_{\pi}\mathbb{E}[R_t | s_t = s, a_t = a, \pi]}$, is the maximum discounted expected reward obtained after executing action $a$ in state $s$. $\pi$ --- the \emph{policy} --- is a mapping from states to actions. 
In Q-learning, the optimal action value function is estimated using the Bellman equation, ${Q^{*}(s,a) = \mathop{\mathbb{E}}_{s'\sim T(s'|s,a)}[r + \gamma \max_{a'} Q^{*}(s',a')| s_t = s, a_t = a]}$, where $T(s'|s,a)$ refers to the state transition distribution. 

RL has previously been used in healthcare settings. Sepsis treatment strategy was tackled by \cite{komorowski}, where models with discretized state and action-spaces were used to deduce treatment policies for septic patients. This work used value-iteration techniques to find an optimal policy \citep{sutton}. The optimal policy was evaluated by directly comparing the Q-values that would have been obtained while following it to those obtained when following the physician's policy. In this work, we build on this research by using continuous state-space models, deep reinforcement learning, and a clinically guided reward function. We extend our qualitative evaluation of the learned policies by examining how policies treat patients of different severity.
\section{Methods}
\subsection{Data and preprocessing}
Data for our cohort were obtained from the Multiparameter Intelligent Monitoring in Intensive Care (MIMIC-III v1.4) database \citep{mimic}; we focused on patients fulfilling the Sepsis-3 criteria \citep{sepsis3} (17,898 in total). For each patient, we have relevant physiological parameters including demographics, lab values, vital signs, and intake/output events. Data are aggregated into windows of 4 hours, with the mean or sum being recorded (as appropriate) when several data points were present in one window. This yielded a $48\times1$ feature vector for each patient at each timestep, which is the state $s_t$ in the underlying MDP. See Section \ref{sec:appendix_features} for more information. 
\subsection{Actions and rewards}
Our action space in this work is discrete. We define a $5\times5$ action space for the medical interventions covering the space of intravenous (IV) fluid (volume adjusted for fluid tonicity) and maximum vasopressor (VP) dosage in a given 4 hour window. We discretized the action space into per-drug quartiles based on all non-zero dosages of the two drugs, and converted each drug at every timestep into an integer representing its quartile bin. We included a special case of no drug given as bin 0. This created an action representation of interventions as tuples of (total IV in, max VP in) at each time. We choose to focus on this action space given the uncertainty in clinical literature of how to adjust these interventions on a per-patient basis, and also their crucial impact on a patient's eventual outcome \citep{waechter2014interaction}. 

The reward function is clinically guided. We use the SOFA score (measuring organ failure) and the patient's lactate levels (measure of cell-hypoxia that is higher in septic patients) as indicators of a patient's overall health. Our reward function penalizes high SOFA scores and increases in SOFA score and lactate levels. Positive rewards are issued for decreases in these metrics. At the terminal timestep of a patient's trajectory, we assign a reward that is positive in the case of survival, and negative otherwise. See Section \ref{sec:appendix_reward} for more information. 
\subsection{Model architecture}
To learn treatment policies, we use neural networks to approximate $Q^{*}(s,a)$, the optimal action value function. 
Our model is based on a variant of Deep Q Networks \citep{atari} which optimizes parameters to minimize a squared error loss between the output of the network, $Q(s,a;\theta)$, and the desired target, $Q_{\textit{target}} = r + \gamma \max_{a'}Q(s',a';\theta)$, observing tuples of the form $<s,a,r,s'>$. The network has a $Q$ output for for all actions. In practice, the expected loss is minimized via stochastic batch gradient descent. However, non-stationarity of the target values can lead to instability, and using a separate network to determine the target Q values ($Q(s',a')$), which is periodically updated towards the main network (which is used to estimate $Q(s,a)$) helps to improve performance. 

Simple Q-Networks have several shortcomings, so we made  modifications to improve our model. Firstly, Q-values are frequently overestimated, leading to incorrect predictions and poor policies. We solve this problem with a Double-Deep Q Network \citep{ddqn}, where the target Q values are determined using actions found through a feed-forward pass on the main network, as opposed to being determined directly from the target network. When finding optimal treatments, we want to separate the influence on Q-values of 1) a patient's \textit{underlying state} being good (e.g. near discharge), and 2) the correct action being taken at that time-step. To this end, we use a Dueling Q Network \citep{dueling}, where the action-value function for a given $(s,a)$ pair, $Q(s,a)$, is split into separate  \emph{value} and \emph{advantage} streams, where the \emph{value} represents the quality of the current state, and the \emph{advantage} represents the quality of the chosen action. We use Prioritized Experience Replay \citep{per} to accelerate learning by sampling a transition from the training set with probability proportional to the previous error. The final network is a fully-connected Dueling Double-Deep Q Network with two hidden layers of size 128, combining the above ideas. For more detail about the architecture, refer to Section \ref{sec:appendix_model}. After training, we obtain the optimal policy for a given patient state as $\pi^{\ast}(s) = \argmax_{a} Q(s,a)$.

\section{Results}
We present qualitative results for the performance of the model in different regimes; that is, timesteps at which patients had relatively low SOFA scores (under 5), timesteps at which patients had medium SOFA scores (5 -- 15) and timesteps at which patients had high SOFA scores (over 15); this is to understand how the model performs on different severity subcohorts.
\vspace{-0.4cm}
\begin{figure}[ht]
 \centering 
 \centerline{\includegraphics[width=5.8in]{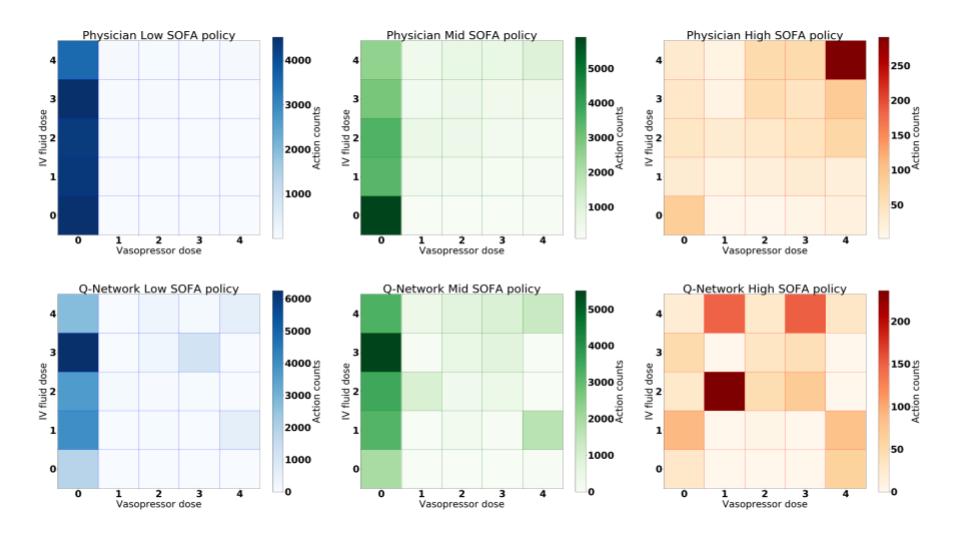}} 
 \caption{Policies learned by the different models, as a 2D histogram, where we aggregate all actions selected by the physician and model on the test set over all relevant timesteps. The axes labels index the discretized action space, where 0 represents no drug given, and 4 the maximum of that particular drug. The model learn to prescribe vasopressors sparingly, a key feature of the physician's policy.}
 \label{fig:pols} 
\end{figure} 

\begin{figure}[h]
 \centering 
 \centerline{\includegraphics[width=6in]{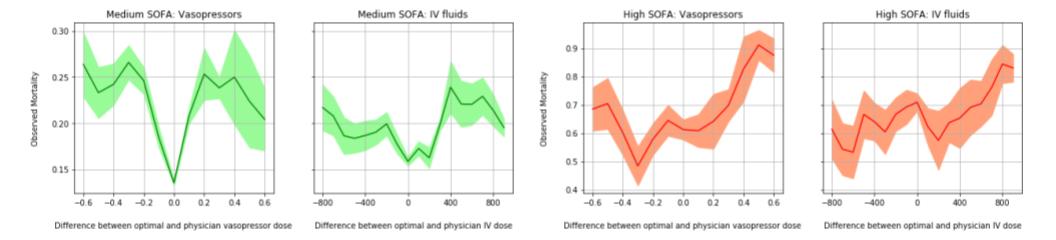}}
 \caption{Comparison of how observed mortality (y-axis) varies with the difference between the dosages recommended by the optimal policy and the dosages administered by clinicians (x-axis) on a held-out test set. For every timestep, this difference was calculated and associated with whether the patient survived or died in the hospital, allowing the computation of observed mortality. We see low mortality with medium SOFA scores for when the difference is zero, indicating that when the physician acts according to the learned policy in this regime we observe more patient survival.}
 \label{fig:diff_mort} 
\end{figure}

Figure \ref{fig:pols} depicts the treatment policies followed by physicians and the model for the three different subcohorts. The action numbers index the different discrete actions selected at a given timestep, and the charts shown aggregate actions taken over all timesteps for those cohorts. Action 0 refers to no drugs given to the patient at that timestep, and increasing actions refer to higher drug dosages, where drug dosages are represented by quartiles. As shown, physicians do not often prescribe vasopressors to patients, unless patients have very high SOFA scores (note the high density of actions corresponding to vasopressor dose = 0 in the first two histograms) and this behavior is reflected in the policy learned by the model. This result is clinically interpretable; even though vasopressors are commonly used in the ICU to elevate mean arterial pressure, many patients with sepsis are not hypotensive and therefore do not need vasopressors. In addition, there have been few controlled clinical trials that have documented improved outcomes from their use~\citep{mullner2004vasopressors}. The discrepancy between the policies learned in the case of high SOFA scores is interesting; the reason for this is likely that only very few data points are available in this regime, and so the policy learned by the Q-network does not mirror that of the physician. This could suggest that it might not be reasonable to use this model for decision support in this regime: an idea that is supported by Figure \ref{fig:diff_mort}.

Figure \ref{fig:diff_mort} shows the correlation between 1) the observed mortality, and 2) the difference between the optimal doses suggested by the policy, and the actual doses given by clinicians. The dosage differences at individual timesteps were binned, and mortality counts were aggregated. The results shown are for timesteps with medium SOFA scores and high SOFA scores; the results for low SOFA scores are not as informative because baseline mortality in this regime is low, and there does not appear to be significant variation in mortality with the dosage of drug given. 
In the case of medium SOFA scores, we observe low mortalities when the optimal dosage and true dosage coincide, i.e. at a difference of 0, indicating the potential validity of the learned policy. The observed mortality proportion then increases as the difference between the optimal dosage and the true dosage increases. 
For high SOFA scores, the model's performance appears to be weaker, with zero deviation not indicating the best outcomes. This could be due to the lack of data points available for this regime, and also the difficulty in learning a strong policy when patients are severely septic; relying on the model's policy might not be recommended for patients with high SOFA scores. Through qualitative analysis we can understand in which regions we may be able to trust the model (and in which regions we cannot), which is essential for using such systems in safety-critical environments.

In addition to qualitative analyses, we can also use Off-Policy Evaluation techniques to estimate the value of our policy, if it were to be deployed. We use the method described by \cite{off-policy-eval}, but note that the value estimates may not be reliable (refer to Section \ref{sec:OPE} for details). We focus on qualitative analyses to  give interpretable insights into our learned policy's efficacy. 

\section{Conclusions}
In this work, we applied deep reinforcement learning to the problem of deducing optimal medical treatments for patients with sepsis. We examined fully continuous state-space/discretized action-space models to find optimal treatment policies, using Dueling Double-Deep Q Networks to learn an approximation for the optimal action-value function, $Q^{*}(s,a)$. We demonstrated that using continuous state-space modeling found interpretable policies that could improve on treatment policies used by clinicians. As future steps, it is important to examine the learned policies on a per-patient basis and investigate alternative modeling strategies (such as model-based reinforcement learning).

\section{Acknowledgements}
This research was funded in part by the Intel Science and Technology Center for Big Data, the National Library of Medicine Biomedical Informatics Research Training grant 2T15 LM007092-22, NIH National Institute of Biomedical Imaging and Bioengineering (NIBIB) grant R01-EB017205, NIH National Human Genome Research Institute (NHGRI) grant U54-HG007963, Imperial College President's PhD Scholarship, and the UK Engineering and Physical Sciences Research Council. The authors would also like to thank Finale Doshi-Velez, David Sontag, and Sumeetpal Singh for their advice.
\bibliography{nips}
\bibliographystyle{plain}
\clearpage
\section{Appendix}
\subsection{Cohort definition}

Following the latest guidelines, sepsis was defined as a suspected infection (prescription of antibiotics and sampling of bodily fluids for microbiological culture) combined with evidence of organ dysfunction, defined by a Sequential Organ Failure Assessment (SOFA) score greater or equal to 2 \citep{sepsis3}. We assumed a baseline SOFA of zero for all patients. For cohort definition, we respected the temporal criteria for diagnosis of sepsis: when the microbiological sampling occurred first, the antibiotic must have been administered within 72 hours, and when the antibiotic was given first, the microbiological sample must have been collected within 24 hours \citep{sepsis3}. The earliest event defined the onset of sepsis. We excluded patients who received no intravenous fluid, and those with missing data for 8 or more out of the 48 variables. This method yield a cohort of 17,898 patients.

The resulting cohort is described in Table \ref{tab:cohort}.
\begin{table}[htbp]
 \centering 
 \begin{tabular}{|l|l||l|l||l|}\hline
  & \% Female & Mean Age & Hours in ICU & Total Population \\ \hline
   	Survivors & 43.6 & 63.4 & 57.6 & 15,583 \\ \hline
	Non-survivors & 47.0 & 69.9 & 58.8 & 2,315 \\ \hline
 \end{tabular}
  \caption{Cohort statistics for subjects that fulfilled the Sepsis-3 criteria.} 
  \label{tab:cohort} 
\end{table}

\subsection{Data extraction}

MIMIC-III v1.4 was queried using pgAdmin 4. Raw data were extracted for all 47 features and processed in Matlab (version 2016b). Data were included from up to 24 hours preceding the diagnosis of sepsis and until 48 hours following the onset of sepsis, in order to capture the early phase of its management including initial resuscitation, which is the time period of interest. The features were converted into multidimensional time series with a time resolution of 4 hours. The outcome of interest was in-hospital mortality.

\subsection{Model Features}
\label{sec:appendix_features}

\textbf{Choice of features}: the included features were chosen to represent the most important parameters clinicians would examine when deciding treatment and dosage for patients; however, some factors not included in our feature vector could serve as confounding factors. Exploring the effect of these is an important future direction. 

The physiological features used in our model are:

\textbf{Demographics/Static}: Shock Index, Elixhauser, SIRS, Gender, Re-admission, GCS - Glasgow Coma Scale, SOFA - Sequential Organ Failure Assessment, Age \newline \newline
\textbf{Lab Values}: Albumin, Arterial pH, Calcium, Glucose, Hemoglobin, Magnesium, PTT - Partial Thromboplastin Time, Potassium, SGPT - Serum Glutamic-Pyruvic Transaminase, Arterial Blood Gas, BUN - Blood Urea Nitrogen, Chloride, Bicarbonate, INR - International Normalized Ratio, Sodium, Arterial Lactate, CO2, Creatinine, Ionised Calcium, PT - Prothrombin Time, Platelets Count, SGOT - Serum Glutamic-Oxaloacetic Transaminase, Total bilirubin, White Blood Cell Count \newline \newline
\textbf{Vital Signs}: Diastolic Blood Pressure, Systolic Blood Pressure, Mean Blood Pressure, PaCO2, PaO2, FiO2, PaO/FiO2 ratio, Respiratory Rate, Temperature (Celsius), Weight (kg), Heart Rate, SpO2 \newline \newline
\textbf{Intake and Output Events}: Fluid Output - 4 hourly period, Total Fluid Output, Mechanical Ventilation \newline \newline
\textbf{Miscellaneous}: Timestep 

\subsection{Reward function}
\label{sec:appendix_reward}
As our objective is to improve patient survival, the model should receive a reward when the patient's state improves, and a penalty when the patient's state deteriorates. This reward should be comprised of the best indicators of patient health; in this situation, these indicators include the patient's SOFA score (which summarizes the extent of a patient's organ failure and thus acts as a proxy for patient health) as well as the patient's lactate levels (a measure of cell-hypoxia that is higher in septic patients because sepsis-induced low blood pressure reduces oxygen perfusion into tissue). An effective reward function should penalize high SOFA scores as well as increases in SOFA score and reward decreases in SOFA scores between states. Similarly, for lactate, increases in lactate should be penalized while decreases in lactate should be rewarded. 

We opted for a reward function for intermediate timesteps as follows:
$$r(s_{\textit{t}}, s_{\textit{t+1}}) = C_{0}\mathbbm{1}(s^{\textnormal{SOFA}}_{\textit{t+1}} = s^{\textnormal{SOFA}}_{\textit{t}}\ \& \  s^{\textnormal{SOFA}}_{\textit{t+1}} > 0) + C_{1}(s^{\textnormal{SOFA}}_{\textit{t+1}}-s^{\textnormal{SOFA}}_{\textit{t}}) + C_{2}\tanh(s^{\textnormal{Lactate}}_{\textit{t+1}} - s^{\textnormal{Lactate}}_{\textit{t}})$$
We experimented with multiple parameters and opted to use $C_{0} = -0.025$, $C_{1} = -0.125$, $C_{2} = -2$. 

At terminal timesteps, we issue a reward of $+15$ if a patient survived their ICU stay, and a negative reward of $-15$ if they did not.

The reason behind the chosen parameters for the intermediate reward, as well as the form of the equation above, was to ensure that the reward was limited in magnitude and would not eclipse the terminal timestep reward at the end of each patient's trajectory. This was also the motivation behind using the $\tanh$ function when dealing with lactate changes: because the maximum change was significantly higher than the average change between timesteps, we opted to use a $\tanh$ to cap the maximum reward/penalty to $|C_{2}|$.

\subsection {Model Architecture and Implementation Details}
\label{sec:appendix_model}
Our final Dueling Double-Deep Q Network network architecture has two hidden layers of size 128, using batch normalization \citep{batchnorm} after each, Leaky-ReLU activation functions, a split into equally sized advantage and value streams, and a projection onto the action-space by combining these two streams together. 

The activation function is mathematically described by: $f(z) = \max(z,0.5z)$, where z is the input to a neuron.
This choice of activation function is motivated by the fact that Q-values can be positive or negative, and standard ReLU, tanh, and sigmoid activations appear to lead to saturation and `dead neurons' in the network. Appropriate feature scaling helped alleviate this problem, as did issuing rewards of $\pm15$ at terminal timesteps to help model stability.

We added a regularization term to the standard Q-network loss that penalized output Q-values which were outside of the allowed thresholds ($Q_{\textnormal{thresh}} = \pm20$), in order to encourage the network to learn a more appropriate Q-function. Clipping the target network outputs to $\pm20$ was also found to be useful. The final loss function was: 
$$\mathcal{L}(\theta) = \mathbb{E}\left[\left(Q_{\textnormal{double-target}} - Q\left(s,a;\theta\right)\right)^{2}\right] + \lambda \cdot \max\left(\left| Q(s,a;\theta)-Q_{\textnormal{thresh}} \right|,0\right)$$ 
$$ Q_{\textnormal{double-target}} = r + \gamma Q(s',\argmax_{a'}Q(s',a';\theta);\theta ')$$
where $\theta$ are the weights used to parameterize the main network, and $\theta '$ are the weights used to parameterize the target network.

We use a train/test split of 80/20 and ensure that a proportionate number of patient outcomes are present in both sets. Batch normalization is used during training. All models were implemented in TensorFlow, with Adam being used for optimization \citep{adam}. 

During training, we sample transitions of the form $<s,a,r,s'>$ from our training set, using the Prioritised Experience replay scheme \citep{per}, perform feed-forward passes on the main and target networks to evaluate the output and loss, and update the weights in the main network via backpropagation. Training was conducted for $80000$ batches, with batch size $32$.

Code is available at \texttt{https://github.com/darkefyre/sepsisrl/}.

\subsection{Off-Policy Evaluation}
\label{sec:OPE}
We use the method of Doubly Robust Off-policy Value Evaluation \citep{off-policy-eval} to evaluate policies. For each trajectory $H$ we compute an unbiased estimate of the value of the learned policy, $V_{\textit{DR}}^{H}$,
, using the recursion $V_{DR}^{H+1-t} = \widehat{V}(s_t) + \rho_t \left(r_t + \gamma V_{DR}^{H-t} -  \widehat{Q}(s_t,a_t)\right)$,
and average the results obtained across the observed trajectories. We can also compute the mean discounted return of chosen actions under the physician policy. See \cite{off-policy-eval} for more information.

This method combines both importance sampling (IS) and approximate Markov Decision Process models (AM) to provide an unbiased, low-variance estimate of the quantity $V_{\textit{DR}}^{H}$.

Initial results indicate that the value estimate of the learned policy $V_{\textit{DR}}^{H}$ is higher than that of clinicians. However, as our learned policy (using the Dueling Double-Deep Q Network) is deterministic, we often rely on only the estimated reward from the AM, which is $\widehat{V}(s_t)$ in the above expression, when estimating values. A deterministic evaluation policy leads to IS terms going to zero if the clinician and learned policy take different actions at any given timestep. We are therefore limited in the accuracy of our value estimates by the accuracy of this estimated reward, and we cannot easily provide statistical guarantees of performance. Improving the quantitative evaluation methodology in this setting is an important direction of future work.
\end{document}